\pdfoutput=1

\documentclass[11pt]{article}

\usepackage{acl}

\usepackage{times}
\usepackage{latexsym}
\usepackage{enumitem}

\usepackage{multirow}
\usepackage{tabularx}
\usepackage[normalem]{ulem}
\usepackage{fnpct}

\usepackage[T1]{fontenc}

\usepackage[utf8]{inputenc}

\usepackage{microtype}

\title{On the Impact of Noises in Crowd-Sourced Data for Speech Translation}

\author{
    Siqi Ouyang\textsuperscript{1}, Rong Ye\textsuperscript{2}, Lei Li\textsuperscript{1} \\ 
    \textsuperscript{1}University of California, Santa Barbara, CA, USA \\
    \texttt{siqiouyang@ucsb.edu, leili@cs.ucsb.edu} \\
    \textsuperscript{2}ByteDance AI Lab, Shanghai, China \\
    \texttt{yerong@bytedance.com}
}
  
\newcommand{\red}[1]{\textcolor{red}{#1}}
\newcommand{\blue}[1]{\textcolor{blue}{#1}}

\begin{document}
\maketitle

\begin{abstract}

Training speech translation (ST) models requires large and high-quality datasets. 
MuST-C is one of the most widely used ST benchmark datasets. It contains around 400 hours of speech-transcript-translation data for each of the eight translation directions. 
This dataset passes several quality-control filters during creation. 
However, we find that MuST-C still suffers from three major quality issues: audio-text misalignment, inaccurate translation, and unnecessary speaker's name. 
What are the impacts of these data quality issues for model development and evaluation? 
In this paper, we propose an automatic method to fix or filter the above quality issues, using English-German (En-De) translation as an example.
Our experiments show that ST models perform better on clean test sets, and the rank of proposed models remains consistent across different test sets. Besides, simply removing misaligned data points from the training set does not lead to a better ST model\footnote{The code is released at \url{https://github.com/owaski/MuST-C-clean}}. 

\end{abstract}
\section{Introduction}


Speech-to-text translation (ST) aims to translate a speech of a certain language into a text translation of another language. Recent advances of end-to-end ST models have been largely boosted by the release of large high-quality parallel datasets \cite{aug_librispeech,mustc,covost2}. A clean test set is essential to evaluate the effectiveness of proposed models, and a sizeable well-aligned training set is important to train powerful ST models \cite{wang-etal-2020-curriculum}.

Currently, the most widely-used ST benchmark dataset is MuST-C~\cite{mustc}. It consists of around 400 hours of speech-transcript-translation data from English into eight languages (German, Spanish, French, Italian, Dutch, Portuguese, Romanian, and Russian). MuST-C was built upon English TED Talks, which are often transcribed and translated by voluntary human annotators. A bilingual sentence-level text corpus is firstly constructed based on sentence segmentation and Gargantua alignment tool~\cite{Gargantua}. Then, the transcription is aligned to the corresponding audio tracks using Gentle forced aligner\footnote{\url{https://github.com/lowerquality/gentle}} built on Kaldi ASR toolkit \cite{kaldi}. During alignment, entire talks are discarded if greater than 15\% of words cannot be recognized, and sentences are removed if none of the words was aligned. 


Though MuST-C passed through several quality-control filters, this dataset is still not perfect. Through manual checking, we find three major quality issues in the dataset -- inaccurate translation, audio-text misalignment, and unnecessary speaker's name. 
Along with the three issues identified, more importantly, we are interested in the following questions: Do they affect the robustness of end-to-end speech translation models trained on this corpus? Can we trust the results from existing works using this data?

In order to answer the above questions, we propose an automatic method to filter or fix the aforementioned errors in both the training and test sets. And based on the original and the fixed datasets, we evaluate many popular ST systems including codebases such as ESPnet~\cite{espnet_st} and published models such as XSTNet~\cite{xstnet}. Our experiments have shown that the performance of models we test is actually better than we thought, and their rank remains consistent across test sets. Besides, simply removing those data points with audio-text misalignment from the training set cannot significantly improve ST models.



\begin{table*}[t!]
    \setlength{\belowcaptionskip}{-0.3cm}
    \small
    \centering
    \begin{tabularx}{\textwidth}{|c|X|X|}\hline
        \textbf{Audio Id} & \textbf{Transcripts} & \textbf{Translations} \\\hline
        ted\_319\_84 & \red{\uline{That's what we were looking forward to.}} That is where we're going — this union, \blue{\dashuline{this convergence of the atomic and the digital.}} & \red{\uline{Danach sehnen wir uns.}} Das ist wo wir hingehen - Diese Einheit, \blue{\dashuline{die Konvergenz des Atomaren und des Digitalen.}}\\\hline
        ted\_319\_85 & \red{\uline{this convergence of the atomic and the digital.}} And so one of the consequences of that, I believe, is that where we have this sort of spectrum of media right now — TV, \blue{\dashuline{film, video — that basically becomes one media platform.}} & \red{\uline{die Konvergenz des Atomaren und des Digitalen.}} Eine Konsequenz davon ist, glaube ich, dass wir dieses aktuelle Spektrum an Medien - TV, \blue{\dashuline{Film, Video - zu einer Medienplatzform wird.}} \\\hline
        ted\_319\_86 & \red{\uline{film, video — that basically becomes one media platform.}} And while there's many differences in some senses, they will share \blue{\dashuline{more and more in common with each other.}} & \red{\uline{Film, Video - zu einer Medienplatzform wird.}} Es wird viele Unterschiede im gewissen Sinn geben, sie werden aber \blue{\dashuline{mehr und mehr miteinander gemeinsam haben.}}\\\hline
    \end{tabularx}
    \caption{\textbf{Examples of misalignment between audio and text.} Extra words that are not in the given transcript but included in the audio are highlighted in \red{\uline{red}}, and missing words that are included in the transcript but not in the audio are highlighted in \blue{\dashuline{blue}}.}
    \label{tab:misalign}
\end{table*}

\section{Quality Issues in MuST-C Corpus}

In this section, we identify three issues that harm the quality of MuST-C dataset. We choose the En-De direction as an example since it is the most widely used direction for demonstrating the performance of ST models.

\noindent\textbf{Audio-Text Misalignment}~
We randomly sample 1000 utterances from the training set of MuST-C En-De dataset and manually verify whether the audio and text are misaligned. We find 69 cases of misalignment out of 1000 given samples. Most of the time, the audio include extra words from the previous or subsequent sentence of its corresponding transcript and translation and omit some of the words of the correct text. This misalignment, once occurs, affects not only one utterance but also utterances around it. 

Table \ref{tab:misalign} shows a typical case where misalignment happens in consecutive utterances. Each audio contains words of its preceding utterance and omits the last few words of its correct text counterpart. Since MuST-C was built by first constructing bilingual text corpus and then aligning English transcripts with audio tracks, audio-translation misalignments usually occur once audio tracks and transcripts are misaligned. In our sample, 68 out of 69 cases follow this observation.
Note that this kind of error can be automatically detected and possibly fixed by a well-trained forced aligner.

\noindent\textbf{Inaccurate Translation}~
We uniformly sample 200 audio-transcript-translation triples from tst-COMMON set and ask human translators proficient in both English and German to label which German translations are not accurate based on given audio files and transcripts.

Table \ref{tab:trans_err} demonstrates typical errors that human translators find.
In the first case, the English word ``unless'' is missing in its German translation, which completely changes the meaning of sentence. In the second case, the German word ``Vollmachtszertifikat'' means ``power of attorney'' rather than ``certificate authority''. In the third case, ``the most peaceful'' is translated to ``very peaceful''. In the last case, German translation adds an extra sentence ``Bei dem vorigen Beispiel ging es darum, Einzelheiten zu finden'' in the beginning that is not expressed in the audio and transcript.

Some of the errors might be caused by human annotators who volunteered to translate the subtitles for the TED Talk (e.g., case 1,2 and 3), and others might be caused by transcript-translation alignment tools used in dataset creation (e.g., case 4). However, it is hard to quantify the number of translation errors, and we will see its empirical impact in the next section.

\begin{table*}[t]
    \setlength{\belowcaptionskip}{-0.5cm}
    \small
    \centering
    \begin{tabularx}{\textwidth}{|c|X|X|}\hline
        \textbf{\#Case} & \textbf{Transcripts} & \textbf{Inaccurate Translations} \\\hline
        I & Woman: 80's revival meets skater-punk, unless it's laundry day. & Frau: 80er Revival trifft auf Skaterpunk, \red{\uwave{\sout{es sei denn,}}} außer am Waschtag. \\\hline
        II& DigiNotar is a certificate authority from the Netherlands -- or actually, it was. & DigiNotar ist ein \red{\uwave{Vollmachtszertifikat}} aus den Niederlanden – bzw. war es das. \\\hline
        III & Steve Pinker has showed us that, in fact, we're living during the most peaceful time ever in human history. & Steve Pinker hat uns gezeigt, dass wir derzeit in einer \red{\uwave{sehr friedlichen}} Zeit der Menschengeschichte leben. \\\hline
        VI & But what if you want to see brush strokes? & \red{\uwave{Bei dem vorigen Beispiel ging es darum, Einzelheiten zu finden,}} aber was, wenn man die Pinselstriche sehen will? \\\hline
    \end{tabularx}
    \caption{\textbf{Examples of inaccurate translations found by human translators.} Errors are highlighted in \red{\uwave{red}}. The strikethrough corresponds to words that are missed in the inaccurate translation.}
    \label{tab:trans_err}
\end{table*}

\noindent\textbf{Unnecessary Speaker's Name}~
Since MuST-C dataset is built on top of subtitles of TED talks, sometimes the subtitle will include additional information like the speaker's name in a multi-speaker scenario. This additional information cannot be recognized given the single audio segment. However, the impact is negligible since names are usually relatively short (less than 20 characters) compared to the entire utterance (more than 100 characters), and it does not frequently happen (around 7\% in our sample). We merely showcase here the existence of such a problem.

To summarize, we have identified three quality issues, misalignment, inaccurate translation, and unnecessary extra information in the MuST-C dataset. In the next section, we will empirically quantify the impact of these issues in training and testing scenarios. 
\section{Examining the Impact of Quality Issues}

In this section we examine the impact of discovered quality issues on both training and test set of MuST-C En-De dataset. We first fix errors for training and test sets. Then we train models on both original and clean training sets and evaluate their empirical performances on test sets with and without errors.

\subsection{Detecting and Fixing Errors}

We apply different techniques to fix training and test sets due to the size difference and different quality requirements. It is unrealistic to fix erroneous translations for the training set since it requires enormous human effort. Thus, we develop an automatic tool to detect the misalignment and remove them to obtain a clean training set. 

Specifically, we first expand the given audio track by one second in both ends and leverage a pretrained automatic speech recognition (ASR) model \cite{w2v2}\footnote{\url{https://huggingface.co/facebook/wav2vec2-large-960h}} to conduct forced alignment between the expanded audio and transcript. If the given alignment exceeds the time range of the original audio by 0.15 seconds, we treat it as a misalignment. However, this alone cannot deal with the case that audio completely covers the transcript but also has extra content. Thus, we use the same model to conduct ASR task to extract the transcript. If the edit distance between the extracted transcript and the transcript given beforehand is larger than 0.7 times length of the given transcript, we also treat it as a misalignment. We choose the hyperparameters based on 1000 random samples of the dataset to achieve a high recall and an acceptable precision (95\% and 82\% measured on these samples), since we want the dataset to be as clean as possible. By removing these misaligned cases, we obtain a clean training set with 194k utterances compared to the original 229k utterances in the MuST-C training set.


For the test set, we uniformly sample 200 data points (about 10\% of tst-COMMON) and manually fix the aforementioned errors one by one. This provides us four versions of test sets:
\begin{itemize}[itemsep=1pt, leftmargin=10pt, parsep=0pt, topsep=1pt]
    \item \textbf{tst-200}: the sampled 200 data points without modification.
    \item \textbf{tst-200-fix-misalignment}: tst-200 with misalignment fixed.
    \item \textbf{tst-200-fix-translation}: tst-200 with translation errors fixed.
    \item \textbf{tst-200-fix-all}: tst-200 with both errors fixed. 
\end{itemize} 
Note that we align the audio tracks and the text translations by adjusting the audio time ranges rather than the translations since misaligned audio tracks correspond to incomplete sentences.

\begin{table*}[!ht]
    \setlength{\belowcaptionskip}{-0.5cm}
    \centering
    \begin{tabular}{l|cc|c}\hline
        \textbf{Models} & \quad \textbf{tst-200} & \textbf{tst-200-fix-all} & \textbf{tst-COMMON} \\\hline
        \multicolumn{4}{c}{\textit{w/o external MT data}} \\\hline
        ESPnet ST & \quad~21.7 & 23.8 & 22.9 \\
        Fairseq ST & \quad~22.4 & 24.3 & 22.7 \\ 
        NeurST & \quad~21.0 & 24.0 & 22.8 \\
        Speechformer & \quad~24.4 & 27.1 & 23.6 \\
        XSTNet base & \quad~25.5 & 27.4 & 25.5 \\\hline
        \multicolumn{4}{c}{\textit{w/ external MT data}} \\\hline
        Baseline & \quad~25.1 & 27.3 & 24.6 \\
        JT-S-MT & \quad~26.0 & 28.4 & 26.8 \\
        XSTNet expand & \quad~28.1 & 30.8 & 27.1 \\
        Chimera & \quad~28.2 & 31.1 & 27.1 \\\hline
    \end{tabular}
    \caption{Empirical performance of models evaluated on different test sets. tst-200 is an uniformly sampled 200-data-point subset of tst-COMMON. tst-200-fix-all is another version of tst-200 with all quality issues fixed.}
    \label{tab:testbleu}
\end{table*}

\subsection{Examining the Impact}

\paragraph{Experiment Setup}
We adopt a baseline model architecture W2V2-Transformer as in \citet{xstnet} which concatenates a pretrained Wav2vec2 audio encoder\footnote{We adopt the wav2vec 2.0 base model, which passes raw waveform through 7 convolution layers and 12 Transformer encoder layers. It can be accessed here \url{https://dl.fbaipublicfiles.com/fairseq/wav2vec/wav2vec_small.pt}} and a Transformer \cite{transformer} with six encoder and decoder layers respectively. We also adopt the same training procedure as \citet{xstnet} except that we also pre-train the Transformer on WMT14 En-De MT dataset. Training arguments can be referred in the Appendix. We have also collected several representative open-sourced models, including codebases (ESPnet \cite{espnet_st}, Fairseq ST \cite{fairseq}, NeurST \cite{neurst}) and published models (JT-S-MT \cite{stjoint}, Chimera \cite{chimera}, XSTNet \cite{xstnet} and Speechformer \cite{speechformer}), to robustify our experiments. The models are tested on the aforementioned four versions of test sets. We report case-sensitive detokenized BLEU scores using sacreBLEU\footnote{\url{https://github.com/mjpost/sacrebleu}}\footnote{BLEU signature: nrefs:1|bs:1000|seed:12345|case:mixed|
eff:no|tok:13a| smooth:exp|version:2.0.0}.

\paragraph{Impact on Model Evaluation}

We are interested in whether the original test set is enough to serve as the metric for offline speech translation. Therefore, we examine if the rank of existing models will be different after fixing the errors. Results are shown in Table \ref{tab:testbleu}.

The BLEU score increase after switching to the clean test set is consistent across all models, indicating that the performance of these models is better than we previously thought. More importantly, the rank of models evaluated on tst-200 is also consistent with that evaluated on tst-200-fix-all. This demonstrates that the original test set, though noisy, can still assess models' performance. 

We also conduct a case study to qualitatively examine the effect after fixing each of the errors. We run Chimera on both misaligned and aligned inputs to evaluate the effectiveness of fixing misalignment. Table~\ref{tab:case_mismatch} shows two cases. As highlighted in \blue{\uline{blue}}, the translations generated by Chimera are more accurate given aligned inputs. 

We also compare the BLEU score difference brought by fixing inaccurate references in Table~\ref{tab:case_trans}. In both cases, the BLEU scores increase by a large margin, indicating the model performs actually better than we originally thought.

\begin{table*}[t]
    \centering
    \begin{tabularx}{\textwidth}{|c|X|X|X|X|}\hline
        \textbf{\#Case} & \textbf{Transcript} & \textbf{Reference} & \textbf{Translation} & \textbf{Translation} \\
         & & &\textbf{w/ Misalignment} & \textbf{w/o Misalignment} \\\hline
        I & Who are they actually supposed to be informing? & Wen wollen Sie eigentlich damit informieren? & Angenommen, wer sind sie eigentlich? & CA: Wer sollen sie eigentlich \blue{\uline{informieren}}? \\\hline
        II & And so if we think about that, we have an interesting situation in hands. & Und deshalb, falls wir darüber nachdenken haben wir eine interessante Situation vor uns. & Wenn wir also darüber nachdenken, haben wir eine interessante Situation. & Wenn wir also darüber nachdenken, haben wir eine interessante Situation \blue{\uline{in unseren Händen}}.\\\hline
    \end{tabularx}
    \caption{Examples of translation with misaligned and without misaligned audio tracks. Improvements brought by aligned inputs are underlined in \blue{\uline{blue}}.}
    \label{tab:case_mismatch}
\end{table*}

\begin{table*}[t]
    \centering
    \begin{tabularx}{\textwidth}{|c|X|X|X|X|X|}\hline
        \textbf{\#Case} & \textbf{Transcript} & \textbf{Inaccurate} & \textbf{Fixed} & \textbf{Translation} & \textbf{BLEU} \\
         & & \textbf{Reference} & \textbf{Reference} & & \\\hline
        I & Steve Pinker has showed us that, in fact, we’re living during the most peaceful time ever in human history. & Steve Pinker hat uns gezeigt, dass wir derzeit in einer sehr friedlichen Zeit der Menschengeschichte leben. & Steve Pinker hat uns gezeigt, dass wir in der Tat in der friedlichsten Zeit der Menschheitsgeschichte leben. & Steve Pinker zeigte uns, dass wir in der Tat in einer der friedlichsten Zeiten der Menschheitsgeschichte leben. & 13.1 $\rightarrow$ 50.7 \\\hline
        II & This idea of fireflies in a jar, for some reason, was always really exciting to me. & Glühwürmchen in einem Glas fand ich immer ganz aufregend. & Die Vorstellung von Glühwürmchen in einem Glas fand ich aus irgendeinem Grund immer ganz aufregend. & Die Idee von Glühwürmchen und einem Kiefer war aus irgendeinem Grund immer sehr aufregend für mich. & 1.6 $\rightarrow$ 19.3 \\\hline
    \end{tabularx}
    \caption{Examples of BLEU score difference brought by fixing inaccurate translations.}
    \label{tab:case_trans}
\end{table*}

\paragraph{Impact on Model Training}
We examine the impact of discovered quality issues on the training set by training baseline models on the original and clean versions of the training set and evaluate them on four versions of test set. The BLEU scores are shown in Table \ref{tab:trainbleu}.

\begin{table}[t]
    \setlength{\belowcaptionskip}{-0.5cm}
    \centering
    \begin{tabular}{l|cc}\hline
        \textbf{Test-set $\backslash$ Train-set} & \textbf{Original} & \textbf{Clean} \\\hline
        tst-200 & 25.06 & 24.38 \\
        tst-200-fix-misalignment & 25.38 & 24.63 \\
        tst-200-fix-translation & 26.86 & 26.99 \\
        tst-200-fix-all & 27.34 & 27.32 \\\hline
        tst-COMMON & 24.60 & 24.03 \\\hline
    \end{tabular}
    \caption{BLEU scores of baseline model trained on raw/clean datasets and evaluated on different test sets.}
    \label{tab:trainbleu}
\end{table}

When tested on tst-200, the baseline model trained using the original training set performs better than the one trained using a clean counterpart. This phenomenon can be attributed to the larger dataset size and similarity between original training set and tst-200. Both scores increase after fixing misalignment and translation. Interestingly, fixing misalignment does not bring higher score increase for the model trained on clean data. After fixing all the errors, both models behave equally well. Based on these results, we conclude that simply removing the misaligned cases in the training set does not positively impact the model.

\section{Related Works}

The quality control of ST datasets is an essential but hard to solve task for dataset creators. MuST-C \cite{mustc} was built upon TED Talks, which naturally comes with the question of inaccurate audio segmentation and audio-text alignment. Other datasets like CoVoST 2 \cite{aug_librispeech,covost2}, which was built by reading given sentences, do not possess this kind of problems. Besides, MuST-C used Gentle to conduct the forced alignment and there are other newly developed forced aligners we can use such as the one we developed in this paper and Montreal Forced Aligner \cite{mcauliffe17_interspeech} which both take advantage of deep Transformer model and large audio datasets.
\section{Conclusion}

In this paper, we first identify three types of error in MuST-C En-De dataset: inaccurate translation, audio-text misalignment, and unnecessary speaker's name. We then examine the impact of these errors by training models on both original and clean datasets and evaluate them on test sets before and after fixing these errors. Empirical results demonstrate that the existing noisy test set can still serve as the metric for evaluating speech translation models. However, the model's performance is actually better than we previously thought. As for training, a clean training set does not significantly benefit the model's performance. 
\section*{Acknowledgements}

Siqi Ouyang is supported by UCSB-IEE-Meta Collaborative Research Grant on AI.

\bibliography{custom}
\bibliographystyle{acl_natbib}

\appendix
\section{Appendix}

\subsection{Training Arguments of W2V2-Transformer}


We first pre-train Transformer on WMT14 En-De MT dataset using Adam optimizer with $\beta_1=0.9,\beta_2=0.98$ and learning rate 5e-4. The effective batch size is 32,768 tokens. We firsly warmup the learning rate by 4k steps and then apply an inverse square root schedule algorithm to it. The norm of gradient is clipped to 10. We set label smoothing to 0.1. The model is trained for up to 500k steps, and we select the one with the highest BLEU score on the validation set. 

Then W2V2-Transformer is fine-tuned on MuST-C En-De dataset. The learning rate is 2e-4 and we warmup the it by 25k steps. The effective batch size is 16M frames. Other hyperparameters are the same as MT pre-training.

\end{document}